\pdfoutput=1

\documentclass{article} % For LaTeX2e
\usepackage{nips10submit_e,times}
\usepackage[latin1]{inputenc}
\usepackage{amsmath}
\usepackage{amsfonts}
\usepackage{amssymb}
\usepackage{graphicx}
\usepackage{subfig}
\usepackage{algorithm}
\usepackage{algpseudocode}
\usepackage[left = 2.5cm,right = 2.5cm,top = 2.7cm,bottom = 2.3cm,a4paper]{geometry}
\newcommand{\argmax}{\operatornamewithlimits{argmax}}

\title{\textbf{Background Subtraction Using the Factored 3-Way \\ Restricted Boltzmann Machines}}

\author{Soonam Lee \qquad Daekeun Kim\\
Department of Electrical Engineering and Computer Science\\
University of Michigan\\
Ann Arbor, MI 48105 \\
\texttt{snlee@umich.edu}\qquad\texttt{housekdk@umich.edu}
}

\nipsfinalcopy % Uncomment for camera-ready version

\begin{document}

\maketitle

%%%%%%%%%%%%%%%%%%%%%%%%%%%%%%%%%%%%%%%%%%%%%%%%%%%%%%%%%%
\begin{abstract}
In this paper, we proposed a method for reconstructing the 3D model based on continuous sensory input. The robot can draw on extremely large data from the real world using various sensors. However, the sensory inputs are usually too noisy and high-dimensional data. It is very difficult and time consuming for robot to process using such raw data when the robot tries to construct 3D model. Hence, there needs to be a method that can extract useful information from such sensory inputs. To address this problem our method utilizes the concept of Object Semantic Hierarchy (OSH). Different from the previous work that used this hierarchy framework, we extract the motion information using the Deep Belief Network technique instead of applying classical computer vision approaches. We have trained on two large sets of random dot images (10,000) which are translated and rotated, respectively, and have successfully extracted several bases that explain the translation and rotation motion. Based on this translation and rotation bases, background subtraction have become possible using Object Semantic Hierarchy.
\end{abstract}

%%%%%%%%%%%%%%%%%%%%%%%%%%%%%%%%%%%%%%%%%%%%%%%%%%%%%%%%%%
\section{Introduction}
\label{sec:intro}

In order to drive the real world in terms of robot's perspective, recognizing the $3$-dimensional environment is one of the most important problem to address. Robot can learn the environment via various sensors such as vision, laser or pressure. We choose to work with vision sensor because it can provide abundant information compared with other sensors. Input data from vision sensor contains not only edge, corner, and line, but also RGB information that helps robot perceive object in its surroundings. Since the input given to the robot is continuous, we use sequence of images as input data.

Although input data contains the whole information about the environment including background and foreground, robot cannot extract any useful information unless the data is processed manually. Thus, the method that filters information from given input data is necessary. In this paper, we proposed a method that helps robot learn the motion of dynamic objects. In order to address this problem, in \cite{bib:Xu1} and \cite{bib:Xu2} introduced a method called the Object Semantic Hierarchy (OSH), which consists of multiple representations with different ontologies. The OSH by itself can provide analysis of given sequential image. Through this analysis, robot can detect dynamic objects and recognize its existence.

However, the OSH is only the framework that helps segmenting the background information. This framework requires some techniques to pull out the invariant property under the camera motion and detect objects. Previous works using this framework applied SIFT feature and HOG \cite{bib:Xu1} or Homography transformation \cite{bib:Xu2}. Here, we would like to incorporate Deep Belief Network (DBN) model so the robot can not only detect the object, but understand it. Reconstructing the 3D model by combining structure of the OSH to differentiate the noise and DBN technique would give robot a better understanding about its environment. A robot could learn $3$-dimensional environment and recognize the real world as human does using our proposed method.

%%%%%%%%%%%%%%%%%%%%%%%%%%%%%%%%%%%%%%%%%%%%%%%%%%%%%%%%%%
\section{Related Work}
\subsection{Review of Previous Work}
\label{sec:Review}
Since the history of background subtraction or background modeling is long, there are lots of approaches to address this problem. The simplest methods for background subtraction are frame difference between two frames such as median, average, or running average of the previous frames. However, these methods are too sensitive to the threshold as well as they do not cope with multiple model background distributions. A number of probabilistic methods have been proposed for solving this problem such as mixtures of Gaussians \cite{bib:Stauffer}, kernel density estimators \cite{bib:Elgammal}, and mean shift \cite{bib:Comaniciu}. Mixtures of Gaussians consider multiple model background distributions by the accumulation of supporting evidence \cite{bib:Stauffer}. In \cite{bib:Elgammal} the background probability density function is given by the histogram of the most recent pixel values. Mean shift \cite{bib:Comaniciu} is a gradient-ascent method that can detect the multiple model distribution together with their covariance. There are also several papers that contribute to background modeling. In \cite{bib:Sheikh1} a bankground is modeled using the joint spatial color model. They recently described the method that using the RANSAC algorithm and the Markov Random Field, do background subtraction under the freely moving camera \cite{bib:Sheikh2}. Apart from the previous methods since this vision sensor can be considered as a significant role of mobile robotics, running this background subtraction algorithm in real time is a crucial step. In \cite{bib:Chris} the method that adaptively deciding the background model was described and they can do real-time tracking. In addition, statistical approach to solve the background subtraction in a real time was introduced in \cite{bib:Thanarat}.

Different from these approaches mentioned above, our method did not use traditional computer vision methods or assumed the model of background. Instead, our proposed method learn the pattern of the motions using the Restricted Boltzmann Machines (RBM) and obtain the global motion from the sequential images. Once this global motion is identified between frames, we can select the regions that violate the global motion. This violated regions against global motion can be considered as the foreground region as suggested in the OSH \cite{bib:Xu1}.

\subsubsection{Restricted Boltzmann Machines}
\label{sec:RBM}
The Deep Belief Network is cutting edge technique originated from neural network where its motivation is to mimic human's perception process. This technique is probabilistic generative models that are composed of multiple layers of stochastic, latent variables. To get the probabilistic model the RBM imported the concept called energy-based model from thermodynamics. The energy function of the RBM can explain the relationship between visible unit, which is from input images, and hidden unit, which is latent variables. Given visible unit \textbf{x} and hidden units \textbf{h} the energy function of the RBM can be written as follows:

\begin{equation}\label{eq:energy}
E(\textbf{x},\textbf{h}) = -\sum_{i,j} x_i h_j W_{ij} - \sum_i b_i x_i - \sum_j c_j h_j
\end{equation}
where \textbf{b} is the visual biases and \textbf{c} is the hidden biases. Using this energy function in Eq (\ref{eq:energy}), the probability distribution can be derived as such:

\begin{align}
P(\textbf{h}_j=1|\textbf{x}) &= g(\textbf{c}_j + W_j\textbf{x}) \label{eq:posprob} \\
P(\textbf{x}_i=1|\textbf{h}) &= g(\textbf{b}_i + W_i^T\textbf{h}) \label{eq:negprob}.
\end{align}
This $g(\cdot)$ can be a sigmoid function, hyperbolic tangent function or Gaussian function depending on the input data type.

The DBN is well known to have good performance in classification because the DBN can actually learn the meaningful bases from the given input. For example, several experiments were introduced to extract the bases that explain different objects \cite{bib:Honglak}. The DBN is represented as a hierarchal model where each layer contains high level information of its previous layer. In the first hidden layer, the hidden units contain the edge like bases from the objects. In the second hidden layer, there is information about the specific part of the objects. As the depth of hidden layer increases, information that hidden layer contains are the higher level features of objects. In addition, since the DBN technique is one of the machine learning techniques, it involves training and testing stages. Many experiments have been performed using the RBM and they turned out that the performance of unsupervised learning or semi-supervised learning is better than the performance of supervised learning in object detection and classification \cite{bib:Bengio}.

\subsubsection{Higher-Order Restricted Boltzmann Machines}
\label{sec:HigherRBM}
Conventional RBM only focused on the mean intensity of each pixel. However, to capture the correlation of multiple images, the visible unit input images should be multiple too. As shown in Eq (\ref{eq:energy}) and (\ref{eq:probability}), the energy function and probability expressions of the RBM only have one visible unit input. We should use higher order Boltzmann Machines to get more than two inputs. 

One of the drawback to use the $nth$-order Boltzmann machines is that the higher order Boltzmann machines are too complicated to compute the hidden units and cannot be applied to the real time data. More specifically, the weight matrix $W$ that connects visible unit and hidden units is n-dimensional tensor and finding this $W$ requires high computational cost. To reduce the computational cost of this weight matrix, the method that reduced the complexity of weight matrix through factorization is proposed in \cite{bib:Memisevic2}. Moreover, new type of RBM called covariance RBM that can capture the correlation of image is introduced in \cite{bib:Ranzato}. This factorization idea in \cite{bib:Memisevic2} is also applied in \cite{bib:Ranzato}. In particular, the $3$-way factored RBM model is introduced to compute the covariance of input data. By applying this model to still image, their proposed method can capture the self-correlation of given image with duplicated images as two visible units \cite{bib:Ranzato}.

\subsubsection{Object Semantic Hierarchy}
\label{sec:OSH}
\begin{figure}[htbp!]
\vspace{-0.1in}
\centering
\includegraphics[width=0.5\textwidth]{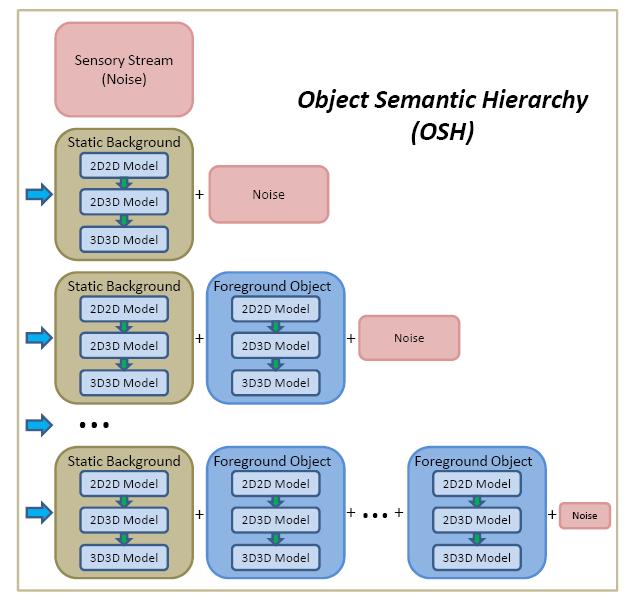}
\caption{The framework of the OSH} \label{fig:OSH}
\vspace{-0.3in}
\end{figure}

In \cite{bib:Xu1, bib:Xu2} the robot initially treats the every input from sensors as noise (Layer 0: Noisy world). Then, using the static background model, the layer can extract the static background (Layer 1: Static Background). During the static background stage a static model of the background is assumed to be time-invariant and dynamic change is considered as noise. Through the $2nd$ layer, the OSH extracts the set of constant 2D object view and the time-variant 2D object poses (Layer 2: 2D object in 2D space). In the $3rd$ layer, it collects each constant 2D image components to reconstruct the time-variant 3D poses (Layer 3: 2D object in 3D space). Finally, in the last layer, the same collection of 2D components but with invariant relations among 3D poses and the time-invariant 3D pose of the objects are extracted as a whole (Layer 4: 3D object in 3D space). 
%Currently, 2D object in 2D space layer is implemented well \cite{bib:Xu1}, but 2D object in 3D space layer is partially succeeded \cite{bib:Xu2} and there is not any starting for 3D object in 3D space layer. 
The framework of the OSH is displayed in Figure \ref{fig:OSH}. In this paper, we utilize the framework of the OSH to perform the background subtraction which is described in the $1st$ stage of the OSH.

\subsection{Main Contribution}
\label{sec:main_contri}
Having mentioned from previous section, the hidden layer can contain the high level features such as edges, the part of objects, and the whole objects from the given image. However, this technique was mainly applied for single still image \cite{bib:Honglak} so far. In \cite{bib:Ranzato} the $3rd$-order RBM that uses two sets of visible units was introduced. Note that previous RBM models use a single set of visible unit. Our intuition in this paper is to apply such a model to extract the motion information of two sequential images, whereas \cite{bib:Ranzato} uses single images as two visible units. More specifically, if two sequential images are used as inputs, the hidden units in the RBM may contain some meaningful information such as the relationship between two images. More precisely, these factors could include each pixel's motion information and correlation rather than edges or objects obtained from using single still image.

Once having the motion information, we can successfully extract optical flow-like field in between two images, namely max-flow field. This is a fairly unconventional but probably quite biologically plausible representation of what happened between two images. There are the number of possible ways of using this flow field including background subtraction. Since the flow field of every pixel is estimated, we could figure out the global motion of each image pairs. Based on the assumption that background regions are large enough compared to foreground regions, we can expect that the global motion regions may represent the background regions. Using the concept of $1st$ stage of the OSH, background subtraction can be done by dividing the two motion subfields: the background motion regions and the motion regions that violate the background motion. 
%So far, nobody have attempted to segment foreground from background using the neural net based model, and thus this research is worthwhile.

%%%%%%%%%%%%%%%%%%%%%%%%%%%%%%%%%%%%%%%%%%%%%%%%%%%%%%%%%%
\section{Technical Content}
\subsection{Summary of the Technical Solution}
\label{sec:summarySol}
As previously discussed, the method using two visible units as inputs was introduced in \cite{bib:Ranzato}. However, different from \cite{bib:Ranzato}, we suggested taking the different images as inputs. More specifically, by taking two sequential images as inputs to the 3-way RBM model, we successfully obtained the motion information between the given inputs. After the motion information from inputs were obtained, we displayed max flow field using the weight matrix computed by the 3-way RBM. Then, based on the max-flow field, we extracted the global motion of two sequential images. Once estimating the flow field of image at pixel level is performed, we can apply the $1st$ stage of the OSH framework to segment foreground object.

\subsection{Details of the Technical Solution}
\subsubsection{Spatial Covariance and Temporal Covariance}
\label{sec:spatCov_tempCov}
Our proposed method starts from the factored 3-way RBM model in \cite{bib:Ranzato}. The energy function of $3rd$-order RBM is expressed as below \cite{bib:Ranzato}:
\begin{equation}\label{eq:3wayRBMenergy1}
E(\textbf{x},\textbf{h}) = -\sum_{i,j,k} x_i x_j h_k W_{ijk} - \sum_j b_j x_j - \sum_k c_k h_k.
\end{equation}
The energy function of the $3rd$-order RBM consists of two visible units $x_i$, $x_j$, one hidden unit $h_k$ and three dimensional tensor $W_{ijk}$. Note that two visible units $x_i$ and $x_j$ are same still image input in \cite{bib:Ranzato}. Using the factorization method, the weight matrix $W_{ijk}$ can be approximately computed by the summation of the products of the two-dimensional weight matrices as:
\begin{equation}\label{eq:wedight_approx}
W_{ijk} \approx -\sum_{f} W_{if} W_{jf} W_{kf}
\end{equation}
where $W_{if}$ represents the weight matrix between visible units and factor and $W_{jf}$ is the weight matrix between another input visible units and factor. Similarly, $W_{kf}$ is the weight matrix between hidden units and factor. Since the $3$-dimensional tensor is factorized by the product of $2$-order weight matrix, the complexity of computation reduces from O($N^3$) to O($N^2$) \cite{bib:Memisevic2}. Moreover, the weight matrix $W_{if}$ and $W_{jf}$ are the same matrix since the duplicated images $x_i$ and $x_j$ are used as visible unit inputs in spatial covariance case \cite{bib:Ranzato}. So given energy function in Eq (\ref{eq:3wayRBMenergy1}) can be simplified as
\begin{equation}\label{eq:3wayRBMenergy2}
E(\textbf{x},\textbf{h}) = -\sum_f(\sum_i x_i W_{if})^2 (\sum_k h_k W_{kf}).
\end{equation}
Using this energy function, we can easily get the probability distribution based on the same inference of conventional RBM such as
\begin{equation}\label{eq:3wayProbability}
P(\textbf{h}_k=1|\textbf{x}) = g(\sum_f W_{kf} (\sum_i x_i W_{if})^2 + c_k). 
\end{equation}

\begin{figure}[htbp!]
	\centering
	\includegraphics[width=0.5\textwidth]{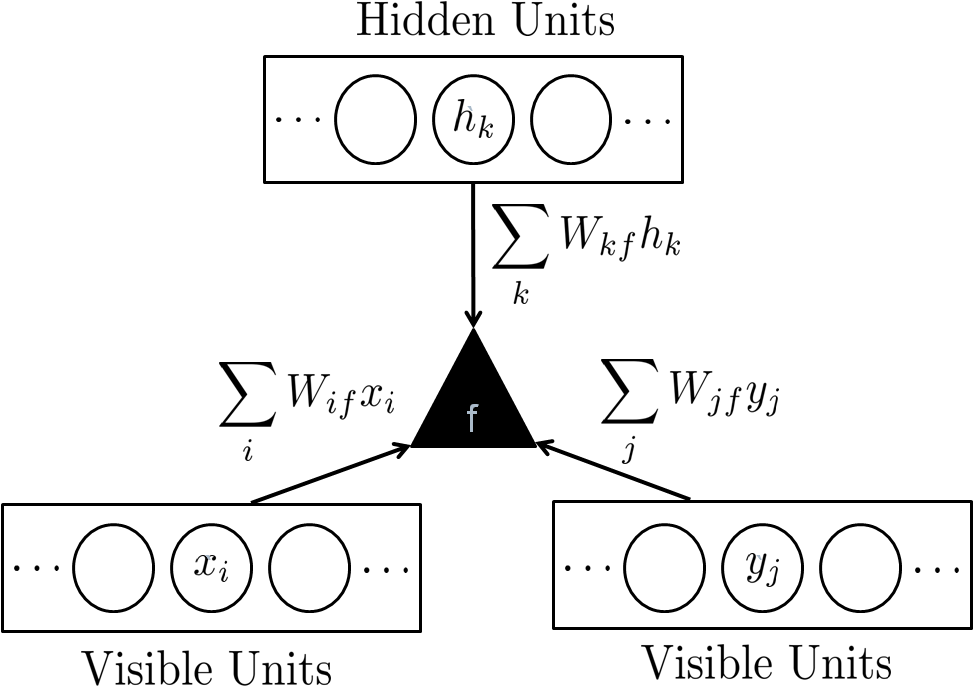}
\caption{The factored $3$-way RBM model in sequential images}
\label{fig:factored_model} % caption for the whole figure
%\vspace{-0.1in}
\end{figure}

Eq (\ref{eq:3wayProbability}) represents the probability distribution based on the energy function shown in Eq (\ref{eq:3wayRBMenergy2}). The relationship between visible units and hidden units can be observed in Figure \ref{fig:factored_model}. In Figure \ref{fig:factored_model}, factored $3$-way RBM model is conceptually presented. Visible units and hidden units are connected with each weight matrix from visible units or hidden units to factor. The weight matrices between visible units and factor ($W_{if}$ and $W_{jf}$) can be computed and obtained by Eq (\ref{eq:3wayRBMenergy2}) and (\ref{eq:3wayProbability}). To get the concrete result, \cite{bib:Ranzato} used the Hybrid Monte Carlo (HMC) method by using leapfrog steps $20$-times per epoch to reject the abnormal output with given probability. Hence, the computed weights based on parameters such as visible units, hidden units converge. 

We expect for these weight matrices to contain the information that can explain the correlation inputs. Since the inputs are the same images in this case, the useful information to explain image would be edge-like features and we denote this method as a spatial covariance method. As shown in Figure \ref{fig:factored_model}, we take the similar concept from 3-way factorized RBM model introduced from \cite{bib:Ranzato}. However, our proposed method uses time sequential images as inputs. We denote our proposed method as a temporal covariance method.

Since our input images are two sequential images denoted as previous image ($x_i$) and current image ($y_j$), respectively, the weight matrix $W_{if}$ between previous image and factor should be different from the matrix $W_{jf}$ between current image and factor. Therefore, the energy functions under the different image inputs are shown as
\begin{equation}
E(\textbf{y},\textbf{h};\textbf{x}) = -\sum_{i,j,k} x_i y_j h_k W_{ijk} - \sum_j b_j y_j - \sum_k c_k h_k.
\end{equation}
Here, $\textbf{x}$ and $\textbf{y}$ are the visible unit inputs and $\textbf{h}$ is hidden units. Different from Eq (\ref{eq:3wayRBMenergy1}), we deal with the conditional energy function to simplify the computational complexity. From this energy function, the probability distributions are similarly computed as Eq (\ref{eq:posprob}) and (\ref{eq:negprob}) such that
\begin{align}
P(h_k=1|\textbf{y}; \textbf{x}) &= g(\sum_{i,j} W_{ijk} x_i y_j +c_k) \label{eq:poshidprobs} \\
P(y_j=1|\textbf{h}; \textbf{x}) &= g(\sum_{i,j} W_{ijk} x_i h_k +b_j) \label{eq:neghidprobs}
\end{align}
where the Eq (\ref{eq:poshidprobs}) represents a positive hidden probability and the Eq (\ref{eq:neghidprobs}) represents a negative hidden probability. Again, this $g(\cdot)$ can be a sigmoid function when the input images are expressed as binary data and the Gaussian distribution when the input images are continuous data.

After computing these hidden probabilities, this algorithm calculates the error between original output ($\textbf{y}$) and new output ($\textbf{y'}$). When we applied the Hybrid Monte Carlo (HMC) technique to our two input images, the reconstruction error failed to reduce and the output results no longer had edge-like features in calculated weights. Therefore, we had to find different method to get the robust weight matrix. We choose to apply the k-steps Contrastive Divergence (CD-k) and the results seem to converge well with only $k = 1$. After the reconstruction error is converged, the reconstruction error is decreased. Brief implementation algorithm is provided from Algorithm 1 at Section \ref{sec:algorithm}.

\subsubsection{Max Flow Field and Background Subtraction}
\label{sec:maxFlowField_backsub}
As we expect to get the edge-like bases from spatial covariance, the hidden units computed by two sequential images have the information of translation and rotation motion behaviors. Once we get the motion bases through the temporal covariance method, max-flow fields are extracted from motion. More specifically, given the motion bases which are factored into three weights, we can estimate optical flow as a binary vector, or as a vector of Bernoulli probabilities. The hidden states inferred from positive hidden probability can give the relationship between the $x_i$ and $y_j$. For every pixel in the $x_i$ the strongest outgoing connection is declared the output pixel which this connection leads to as the target of corresponding input pixel. This is denoted as a max-flow field that maps every input pixel to an output pixel.
% and it is probably the approach that comes closest to a classical flow field.

In order to apply this max-flow field to subtract background, we need to assume two things; the background has a consistent rigid body under the camera motion and the portion of background is dominant from the dynamic objects. Under the these assumptions, finding the global motion can be done by deducing the max-flow field in pixel level. Here, since the robot moves in the real world, robot's camera sensor should not be considered as a fixed camera. Our method should segment foreground objects from background taking consideration of randomly moved camera motion. To subtract background we need to know about the background motion information.

After global motion is estimated, since background region is bigger than foreground regions as we already assumed, the regions that are governed by global motion can be regarded as background regions. Similar to the OSH framework, the regions that violate the invariant property (global motion) will be treated as foreground regions. As such, subtracting the background regions using max-flow field is possible. Brief implementation algorithm is provided from Algorithm \ref{algo:maxFlowField} at Section \ref{sec:algorithm}.

\subsection{Main Implementation Algorithm}
\label{sec:algorithm}

%\pagebreak\newpage
\begin{algorithm}[htbp!]
\caption{Train3WayRBM(\textbf{x},\textbf{y})}\label{algo:3wayRBM}
\begin{algorithmic}[1]
    \State $W_{xf}$ is the input weight matrix, of dimension (number of input units, number of factors)
    \State $W_{yf}$ is the output weight matrix, of dimension (number of output units, number of factors)
    \State $W_{hf}$ is the hidden weight matrix, of dimension (number of hidden units, number of factors)
    \State $\textbf{hbias}$ is the bias vector for hidden units
    \State $\textbf{ybias}$ is the bias vector for output units
    \\
    \For {epoch $ = 1$ to \emph{numEpoch}}
        \For {batch $ = 1$ to \emph{numBatches}}
            \State $\bullet$ Given \textbf{x} and \textbf{y}, compute the positive hidden probability.
            \State $\bullet$ Compute the positive gradient of the energy function with respect to the parameters \State and positive hidden probability.
            \State $\bullet$ Sample from the positive hidden probability to get the hidden units.
            \State $\bullet$ Compute the negative gradient of the energy function with respect to the parameters \State and negative hidden probability.
            \State $\bullet$ Update the parameters by subtracting positive gradient and adding negative gradient.
        \EndFor
    \EndFor
    \State \textbf{return} [$W_{xf}, W_{yf}, W_{hf}, \textbf{hbias}, \textbf{ybias}$]
\end{algorithmic}
\end{algorithm}

We described the brief algorithm of 3-way RBM for learning weights and biases. Implementation is done with Matlab by referring Memisevic's paper \cite{bib:Memisevic2}. In our implementation we used $200$ factors, $100$ mapping units, $0.9$ for momentum, $0.01$ for learning rate, and $0.02$ for target hidden probability as detailed in Algorithm \ref{algo:3wayRBM}. These values were chosen by parameter tuning on a development set of $100$ inference tasks. After training, the model can infer optical flow from a pair of test images with factored weights by computing the hidden unit activations as detailed in Algorithm \ref{algo:maxFlowField}.
The representation of optical flow in the model is as a binary vector and the binary vector over hidden units defines a matrix that maps the input image to the output image. One can find, for example, for every input pixel the strongest outgoing connection to the output pixel that leads to as the target of this input pixel. This defines a max-flow field that maps every input pixel to an output pixel.
\begin{algorithm}[htbp!]
\caption{MaxFlowField(\textbf{x},\textbf{y}, $W_{xf}, W_{yf}, W_{hf}, \textbf{hbias}, \textbf{ybias}$)}\label{algo:maxFlowField}
\begin{algorithmic}[1]
    \State Given a pair of images with parameters, calculate the hidden unit activations.
    \ForAll {input units $i$}
        \State $\bullet$ Compute the input factor units with $ith$ mask.
        \State $\bullet$ Find a reconstructed output units ($\mathbf{\hat{y}}$) through $ith$ mask.
        \State j $\gets$ $\argmax_{i}(\mathbf{\hat{y}})$
        \State $\bullet$ Maps index $i$ and index $j$ into 2D-matrix.
        \State $\bullet$ Make a connection between $i$ and $j$ into 2D-matrix which indicates the optical flow.
    \EndFor
\end{algorithmic}
\end{algorithm}
\vspace{-0.1in}

%%%%%%%%%%%%%%%%%%%%%%%%%%%%%%%%%%%%%%%%%%%%%%%%%%%%%%%%%%
\section{Experiments}

\subsection{Learning Weights}
\label{sec:learning_weights}

As the first stage of experiments, we trained on different types of translations and rotations on dot binary images which were generated randomly with about ten percents of the pixels on. $10,000$ created image of size $13 \times 13$ pixels were used to do the training for translation and rotation and each of their corresponding pairs as a sequential frame was transformed at random direction, respectively. Implementation results of translation and rotation are shown in Figure \ref{fig:translation_bases} and Figure \ref{fig:rotation_bases}, respectively.

Figure \ref{fig:translation_bases} shows the results for finding the temporal covariance between sequential frames when translation motion is trained and Figure \ref{fig:rotation_bases} shows the similar results when the rotation motion is trained. Since the input images for training translations or rotations are the binary inputs, the given probability that shows from the Eq (\ref{eq:poshidprobs}) and (\ref{eq:neghidprobs}) should be the sigmoid function. Note that Eq (\ref{eq:neghidprobs}) should be Gaussian if the input image is the real-valued input. Empirically, the epoch was set to $500$ for optimal results in the translation training and the epoch was set to $700$ for optimal results in the rotation training. As shown in Figure \ref{fig:translation_bases} and \ref{fig:rotation_bases}, corresponding motion information was extracted successfully. From the learned weights we can infer optical flow as well as reconstruct new output images as shown in the next section.

\subsection{Max-Flow Field and Reconstructed Output}
\label{sec:maxFlowField_recon}
\begin{figure}[htbp!]
	\centering
	\subfloat[Left: input $W_{xf}$ bases]
		 {\label{fig:W_xf_translation}\includegraphics[width=0.43\textwidth]{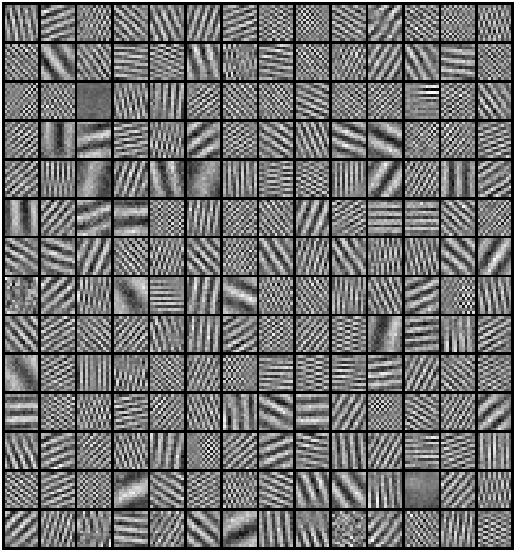}}
    \qquad \qquad
 	\subfloat[Right: output $W_{yf}$ bases]
		 {\label{fig:W_yf_translation}\includegraphics[width=0.43\textwidth]{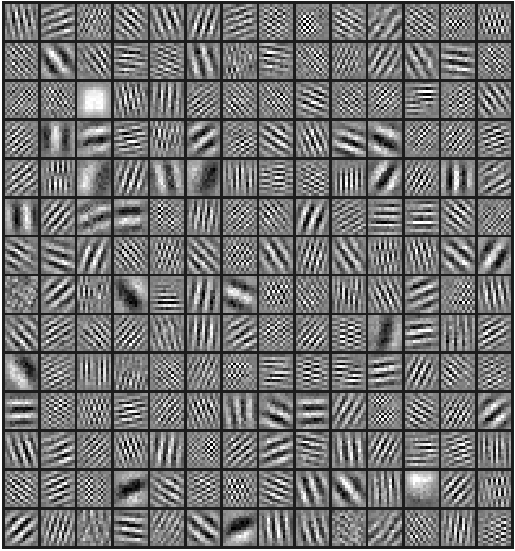}}
\caption{Visible Factor Weights for two sequential input images \textbf{x} and \textbf{y} (Translation). It shows the translation information for each basis.}
\label{fig:translation_bases} % caption for the whole figure
\end{figure}

\begin{figure}[htbp!]
	\centering
	\subfloat[Left: input $W_xf$ bases]
		 {\label{fig:W_xf_rotation}\includegraphics[width=0.43\textwidth]{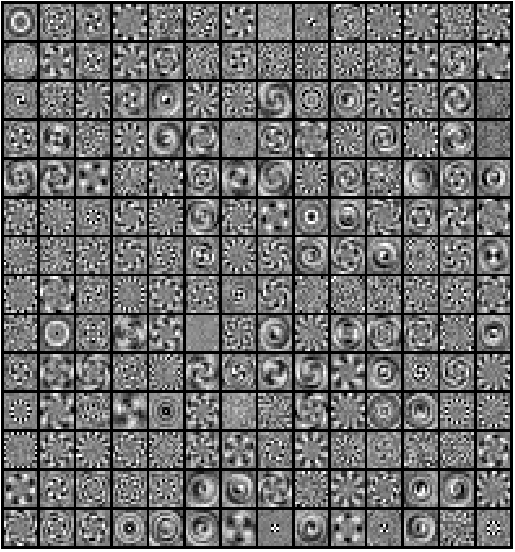}}
    \qquad \qquad
 	\subfloat[Right: output $W_yf$ bases]
		 {\label{fig:W_yf_rotation}\includegraphics[width=0.43\textwidth]{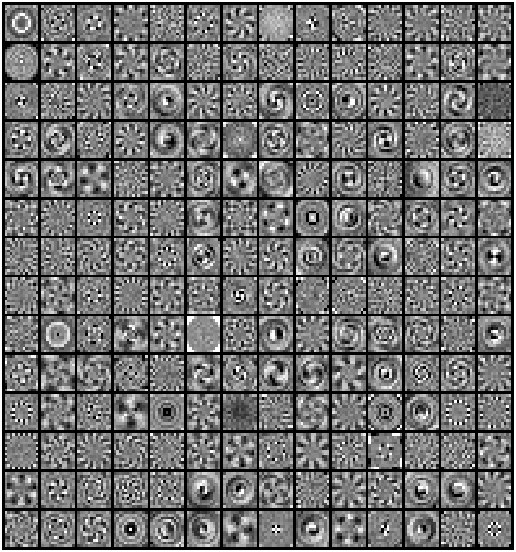}}
\caption{Visible Factor Weights two sequential input images \textbf{x} and \textbf{y} (Rotation). It shows the rotation information for each basis.}
\label{fig:rotation_bases} % caption for the whole figure
\end{figure}

In order to verify whether the model is able to infer transformations such as translation and rotation, we constructed max-flow fields and then displayed them. More specifically, given the learned weights shown in Figure \ref{fig:translation_bases} and \ref{fig:rotation_bases}, we can infer optical flow pattern from pairs of test images by computing hidden unit activations. It is possible to visualize motion flows in the hidden units by drawing arrows from the input pixel to the output pixel. To express the flow field efficiently, we drew only one arrow from the each input pixel which is selected by maximizing the corresponding output pixel location. This connection is defined as a max-flow field as we already discussed in previous section. Note that this process may loose some information since it cannot visualize the case where an input pixel is mapped to multiple output pixels and it may loose all information about uncertainty \cite{bib:Memisevic1}.


\begin{figure}[htbp!]
	\centering
	\subfloat
		 {\includegraphics[width=0.5\textwidth]{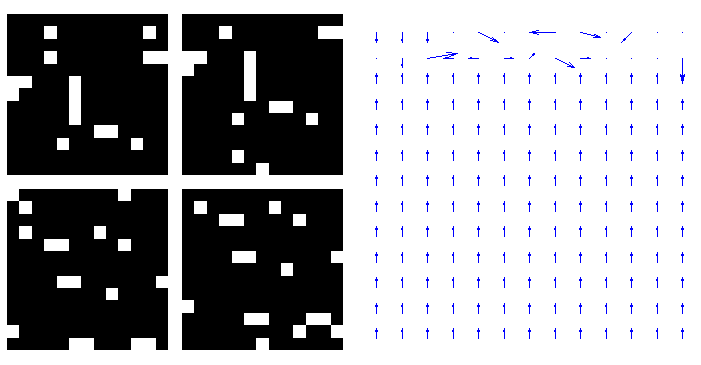}}
 	\subfloat
		 {\includegraphics[width=0.5\textwidth]{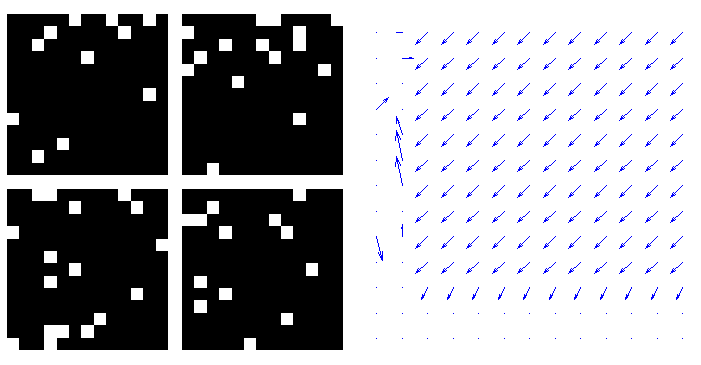}}\\
	\subfloat
		 {\includegraphics[width=0.5\textwidth]{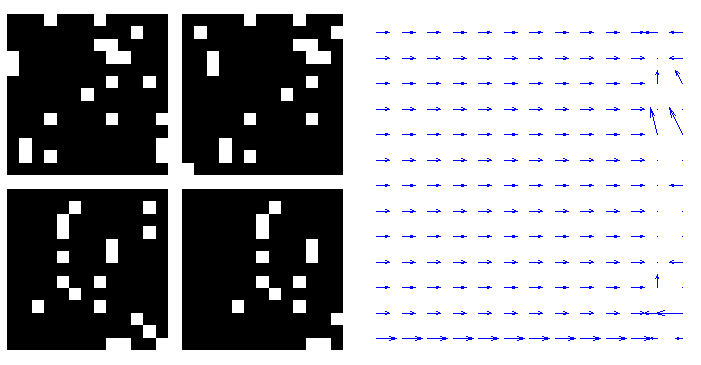}}
 	\subfloat
		 {\includegraphics[width=0.5\textwidth]{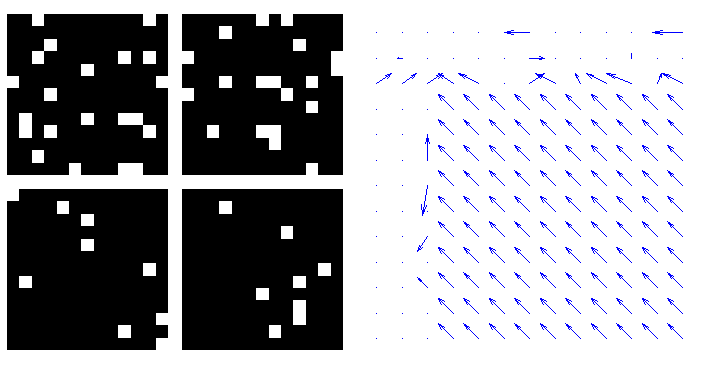}}
\caption{Translation test set. Each group of binary random dot images from left to right represent: input images; output images. The rightmost images indicate representations of inferred max-flow fields between input images and output images} \label{fig:maxflow_translation} % caption for the whole figure
\vspace{-0.2in}
\end{figure}

\begin{figure}[htbp!]
\vspace{-0.1in}
	\subfloat
		 {\includegraphics[width=0.5\textwidth]{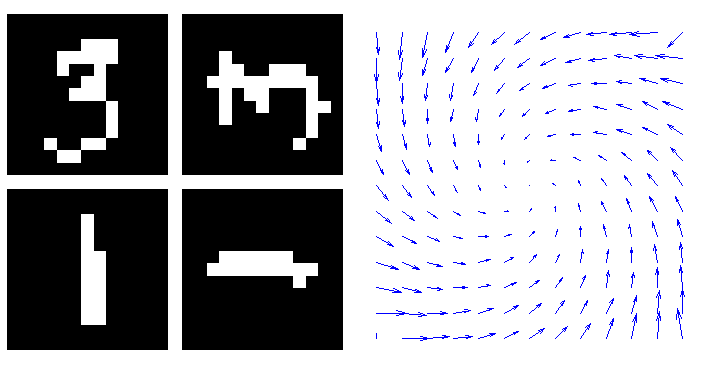}}
 	\subfloat
		 {\includegraphics[width=0.5\textwidth]{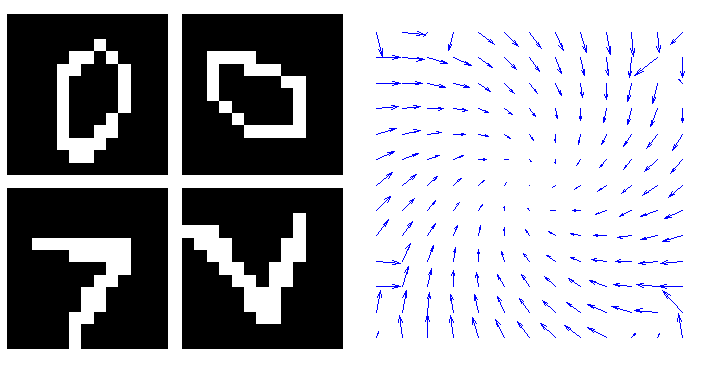}}\\
	\subfloat
		 {\includegraphics[width=0.5\textwidth]{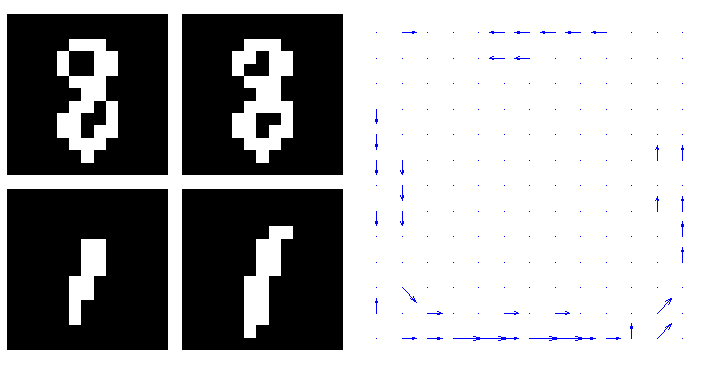}}
 	\subfloat
		 {\includegraphics[width=0.5\textwidth]{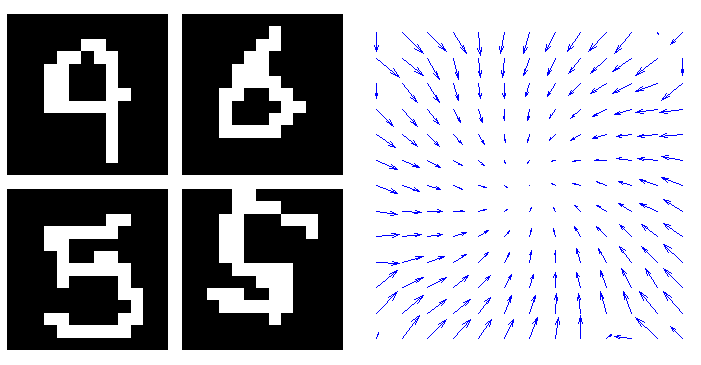}}
\caption{Rotation test set. Each group of binary MNIST images from left to right represent: input images; output images. The rightmost images indicate representation of inferred max-flow field between input images and output images} \label{fig:maxflow_rotation} % caption for the whole figure
\end{figure}

Figure \ref{fig:maxflow_translation} and \ref{fig:maxflow_rotation} shows four image pairs of translation and rotation and each inferred max-flow fields, respectively. For translation, given a test set of $1,000$ binary random dot $13 \times 13$ images, we constructed $9$ kinds of transformations for this experiments such as no shift, up, down, left, right, and four diagonal directions as output images. After generating learned weights, $100$ hidden units are used to map into input images and output images. Hence, we could extract max-flow fields as shown in the rightmost picture for each pair in Figure \ref{fig:maxflow_translation}. For rotation, given a test set of $1,000$ binary MNIST $13 \times 13$ images, we rotated images ranging from $0^{\circ}$ to $360^{\circ}$ as output images. With the same configuration and procedure we could extract max-flow fields as shown in the rightmost picture for each pair in Figure \ref{fig:maxflow_rotation}.

From the max-flow fields, we can easily observe that there is a global motion between two sequential images. As an example of translation shown in Figure \ref{fig:maxflow_translation}, a group of flow fields indicates one dominant direction such as up, lower-left diagonal, right, and upper-left diagonal, respectively. Similarly, as an example of rotation in Figure \ref{fig:maxflow_rotation}, we can clearly observe that there is a global rotation motion. Each input and output pairs represents the global rotation motion such as counterclockwise, clockwise, no-rotation, and reverse turns, respectively.

\vspace{-0.1in}

\begin{figure}[htbp!]
\centering
 {
 \includegraphics[width=1.0\textwidth]{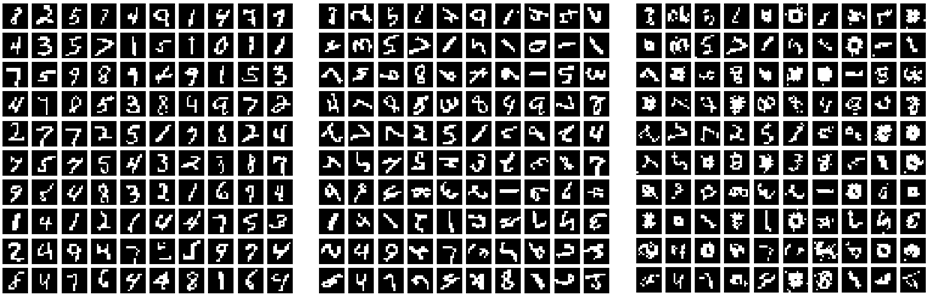}}
 \caption{Analogies in Rotation (Example of images in the MNIST data set). Left column: Input images ($x_i$), Center column: Output images ($y_j$), Right column: Reconstructed images using the input images and inferred max-flow fields which came from hidden unit activations}
\label{fig:MNIST_test}
\vspace{-0.1in}
\end{figure}

Moreover, once we compute the correspondence between input image and output image pairs, we can apply these transformations to previously unseen images. Whatever the input image is, it is successfully reconstruct the output image by analogy since we know the max-flow fields. Figure \ref{fig:MNIST_test} illustrates the reconstructed images.

%%%%%%%%%%%%%%%%%%%%%%%%%%%%%%%%%%%%%%%%%%%%%%%%%%%%%%%%%%
\section{Future Work and Conclusion}
\label{sec:conclusion}
We have proposed a method that combines OSH framework and DBN techniques in order to learn the foreground motion information in a video sequence. Our first step was to reproduce the work of existing DBN technique,called 3-way factored RBM \cite{bib:Ranzato}. There was a further modification to accept two different images as visible units. We have shown that the DBN can not only extract the useful information about single image but also find the correlation between two sequential images, for example, the max-flow fields. With these max-flow fields, we could estimate the global motions.
 
Some of remaining future work includes background and foreground segmentation. As defined in OSH \cite{bib:Xu1}, \cite{bib:Xu2}, we consider the regions that move along to the global motion as background region. On the contrary, foreground region is defined such that the region violates the global motion. Since we finished extracting the flow-field from given test set and estimated the direction of the global motion regions which is become possible to determine background region followed by the foreground regions segmentation. We further believe that building on top of our proposed technique could construct the 3D modeling since we do not need to consider the background information.

%%%%%%%%%%%%%%%%%%%%%%%%%%%%%%%%%%%%%%%%%%%%%%%%%%%%%%%%%%

\end{document}